%% file: main.tex
\documentclass{article}

\usepackage{microtype}
\usepackage{graphicx}
\usepackage{subcaption}
\usepackage{booktabs} 

\usepackage{multirow}

\usepackage{hyperref}

\usepackage[accepted]{icml2026}

\usepackage{amsmath}
\usepackage{amssymb}
\usepackage{mathtools}
\usepackage{amsthm}
\usepackage{xspace}

\makeatletter
\DeclareRobustCommand\onedot{\futurelet\@let@token\@onedot}
\def\@onedot{\ifx\@let@token.\else.\null\fi\xspace}

\def\eg{\emph{e.g}\onedot}

\makeatother

\renewcommand{\paragraph}[1]{\vspace{2pt}\noindent\textbf{#1}}
\usepackage[capitalize,noabbrev]{cleveref}

\theoremstyle{plain}

\theoremstyle{definition}

\theoremstyle{remark}

\usepackage[textsize=tiny]{todonotes}

\definecolor{tticblue}{RGB}{0, 94, 184}
\icmltitlerunning{From Seeing to Thinking: Decoupling Perception and Reasoning Improves Post-Training of Vision-Language Models}

\begin{document}

\twocolumn[
  \icmltitle{From Seeing to Thinking: Decoupling Perception and Reasoning Improves Post-Training of Vision-Language Models}



  \icmlsetsymbol{equal}{*}

  \begin{icmlauthorlist}
    \icmlauthor{Juncheng Wu}{amazon,ucsc}
    \icmlauthor{Hardy Chen}{ucsc}
    \icmlauthor{Haoqin Tu}{ucsc}
    \icmlauthor{Xianfeng Tang}{}
    \icmlauthor{Freda Shi}{waterloo,vector}
    \icmlauthor{Hui Liu}{}
    \icmlauthor{Hanqing Lu}{amazon}
    \icmlauthor{Cihang Xie}{ucsc}
    \icmlauthor{Yuyin Zhou}{ucsc}
    \\
    \textbf{Project Page: }\url{https://ucsc-vlaa.github.io/VLM-CapCurriculum/}
  \end{icmlauthorlist}

  \icmlaffiliation{ucsc}{UC Santa Cruz}
  \icmlaffiliation{amazon}{Amazon}
  \icmlaffiliation{waterloo}{University of Waterloo}
  \icmlaffiliation{vector}{Vector Institute, Canada CIFAR AI Chair}

  \icmlcorrespondingauthor{Juncheng Wu}{jwu418@ucsc.edu}
  \icmlkeywords{Machine Learning, ICML}

  \vskip 0.3in
]



\printAffiliationsAndNotice{}  

\begin{abstract}

Recent advances in vision-language models (VLMs) emphasize long chain-of-thought reasoning; yet, we find that their performance on visual tasks is primarily limited by a lack of visual perception as opposed to reasoning itself. In this work, we systematically study the interplay between perception and reasoning in VLM post-training by decomposing their capabilities into three separate training stages: visual perception, visual reasoning, and textual reasoning, incorporating specialized training data. We demonstrate that visual perception (a) requires targeted optimization with specialized data; (b) serves as a fundamental scaffold that should be solidified through staged training before refining visual reasoning; and (c) is more effectively learned via RL than caption-based SFT. Our experiments across multiple VLMs demonstrate that staged training consistently improves both visual perception and reasoning performance over merged training. Notably, models trained with our approach achieve 1.5\% higher reasoning accuracy with 20.8\% shorter reasoning traces, suggesting that superior perception reduces the need for excessive reasoning. Furthermore, we show that this capability-based staging represents a new curriculum dimension orthogonal to traditional difficulty-based curricula, and combining both yields further additive gains. Our staged-training models achieve superior performance among open-weight VLMs, establishing advanced results on several visual math and perception (\eg, +5.2\% on WeMath and +3.7\% on RealWorldQA) tasks compared with the base counterpart.
\end{abstract}

\input{sections/intro}
\input{sections/related_works}

\input{sections/method}

\input{sections/experiment}

\input{sections/conclusion}








\section*{Impact Statement}
This paper presents work whose goal is to advance the field of Machine
Learning. There are many potential societal consequences of our work, none
of which we feel must be specifically highlighted here.

\newpage
\bibliography{example_paper}
\bibliographystyle{icml2026}

\newpage
\appendix
\onecolumn
\section{Appendix}

\subsection{Detailed Hyperparameter Setting}
\label{app:hyperparameter}

We provide the full hyperparameter in Table~\ref{tab:full_hparams}
For all remaining training parameters not listed in the table, we follow the default settings of EasyR1~\cite{zheng2025easyr1} to ensure a controlled comparison and reproducibility.

\begin{table}[ht]
\centering
\small
\caption{\textbf{Key hyperparameters used in our Stage-3 training.}}
\begin{tabular}{ll}
\hline
\textbf{Hyperparameter} & \textbf{Value} \\
\hline
Max prompt length & 2048 \\
Actor dtype & bf16 \\
Actor optimizer & adamw\_bf16 \\
Actor micro-bsz (update) & 16 \\
Actor micro-bsz (experience) & 64 \\
Offload params / optim & False / False \\
Rollout GPU mem util. & 0.7 \\
Tensor parallel size & 1 \\
Reward type & sequential \\
GPUs per node & 8 \\
\hline
\end{tabular}
\label{tab:full_hparams}
\end{table}

\subsection{More Experimental Results}
\label{app:more_results}

\paragraph{Ablation of each training stage.}
\label{app:staged_ablation}
\begin{table}[ht]
\centering
\small
\caption{\textbf{Ablation study of different staged training combinations on Qwen3-VL-8B (Accuracy \%).}}
\begin{tabular}{lccccccc}
\hline
\textbf{Training Stages} & \textbf{MVista} & \textbf{MVision} & \textbf{WeMath} & \textbf{A-OKVQA} & \textbf{RWQA} & \textbf{POPE} & \textbf{AVG} \\
\hline
Base Model & 72.40 & 26.32 & 50.86 & 87.86 & 70.85 & 88.11 & 66.07 \\
\hline
Stage 3 & 73.90 & 26.64 & 56.10 & 86.03 & 73.59 & 87.69 & 67.33 \\
Stage 1$\rightarrow$3 & 73.90 & 29.28 & 58.76 & 86.55 & 73.59 & 87.55 & 68.27 \\
\hline
Stage 2$\rightarrow$3 & 73.80 & 27.30 & 56.10 & 86.90 & 73.86 & 87.35 & 67.55 \\
Stage 1$\rightarrow$2$\rightarrow$3 & 75.90 & 28.62 & 56.10 & 86.29 & 74.51 & 87.88 & 68.22 \\
\hline
\end{tabular}
\label{tab:stage_ablation_qwen3vl8b}
\end{table}

The ablation results in Table~\ref{tab:stage_ablation_qwen3vl8b} further verify the critical role of visual perception training. 
Compared with applying Stage~3 (visual reasoning) alone, introducing visual perception-oriented Stage~1 before Stage~3 yields clear gains on visual math benchmarks, with MVision improving from 26.64\% to 29.28\% and WeMath from 56.10\% to 58.76\%, and the overall average increasing from 67.33\% to 68.27\%. 
In contrast, directly adding Stage~2 before Stage~3 leads to only marginal changes (AVG: 67.33\% \textit{v.s.} 67.68\%), indicating that reasoning-oriented improvements largely saturate when perception remains weak. 
Moreover, incorporating Stage~1 prior to both Stage~2 and Stage~3 further enhances performance over Stage~2$\rightarrow$3, particularly on MVista (75.90\% \textit{v.s.} 73.80\%). 
Together, these findings demonstrate that visual perception constitutes a dominant bottleneck in current VLMs, and explicitly strengthening perception is a prerequisite for unlocking effective downstream reasoning improvements.

\paragraph{Impact of Training Vision Encoder.}
\label{app:vision_encoder}

\begin{table}[ht]
\centering
\small
\caption{\textbf{Effect of vision encoder freezing strategies under staged and merged training (Accuracy \%). }
\emph{Mixed} denotes the strategy used in the main paper, where the vision encoder is frozen in Stage~2 but unfrozen in Stage~1 and Stage~3.}
\begin{tabular}{llccccccc}
\hline
\textbf{Model / Training} & \textbf{Vision Encoder} & \textbf{MVista} & \textbf{MVision} & \textbf{WeMath} & \textbf{A-OKVQA} & \textbf{RWQA} & \textbf{POPE} & \textbf{AVG} \\
\hline
Qwen2.5-VL-7B / Staged & Mixed & 71.20 & 21.71 & 38.29 & 87.60 & 70.46 & 86.84 & 62.68 \\
Qwen2.5-VL-7B / Staged & All Freeze & 70.00 & 20.72 & 38.38 & 86.90 & 69.02 & 86.33 & 61.89 \\
Qwen2.5-VL-7B / Staged & All Open & 70.50 & 22.37 & 38.67 & 87.16 & 69.93 & 86.08 & 62.45 \\
Qwen2.5-VL-7B / Merged & All Freeze & 70.90 & 20.39 & 36.67 & 85.59 & 70.07 & 84.35 & 61.33 \\
Qwen2.5-VL-7B / Merged & All Open & 70.00 & 20.39 & 37.43 & 86.03 & 69.28 & 84.74 & 61.31 \\
\hline
Qwen3-VL-8B / Staged & Mixed & 75.90 & 28.62 & 56.10 & 86.29 & 74.51 & 87.88 & 68.22 \\
Qwen3-VL-8B / Staged & All Freeze & 75.20 & 30.26 & 56.57 & 86.11 & 76.21 & 87.42 & 68.63 \\
Qwen3-VL-8B / Staged & All Open & 75.30 & 31.91 & 54.67 & 86.55 & 74.90 & 87.18 & 68.42 \\
Qwen3-VL-8B / Merged & All Freeze & 74.30 & 25.99 & 54.38 & 85.33 & 73.86 & 87.40 & 66.88 \\
Qwen3-VL-8B / Merged & All Open & 73.80 & 28.95 & 55.43 & 85.50 & 75.56 & 87.19 & 67.74 \\
\hline
\end{tabular}
\label{tab:vision_encoder_ablation}
\end{table}

Table~\ref{tab:vision_encoder_ablation} compares different vision encoder freezing strategies under both staged and merged training. 
Across settings, varying the vision encoder between fully frozen, fully trainable, and the proposed mixed strategy leads to relatively small performance differences, suggesting that encoder freezing alone is not a dominant factor governing final performance. 
In contrast, staged training consistently outperforms merged training under comparable encoder configurations on both Qwen2.5-VL-7B and Qwen3-VL-8B. 
For example, on Qwen2.5-VL-7B, staged training models achieve higher average accuracy (up to 62.68\%) than their merged counterparts (around 61.3\%), while on Qwen3-VL-8B, staged training reaches 68.22\%--68.63\% compared to 66.88\%--67.74\% under merged training. 
These consistent gains across architectures indicate that the staged training paradigm itself, rather than specific encoder freezing heuristics, is the primary driver of performance improvements.

\paragraph{Exact Values in Section~\ref{sec:perception_data}.}
\label{app:fig2_exact_numbers}

\begin{table}[t]
\centering
\small
\setlength{\tabcolsep}{2.8pt}
\caption{\textbf{Comparison between the base model, the model trained with reasoning-only, and perception+reasoning data (Accuracy \%).} Incorporating perception data improves visual math while maintaining perception capabilities. We show standard error bars in Figure~2, and the exact values are provided in this table.}
\begin{tabular}{lcccccccc}
\hline
\textbf{Model} & \textbf{MVista} & \textbf{MVerse (VI)} & \textbf{WeMath} & \textbf{A-OKVQA} & \textbf{RWQA} & \textbf{MMStar} & \textbf{POPE} & \textbf{AVG} \\
\hline
Qwen2.5-VL-7B & 68.50 & 26.40 & 30.86 & 86.81 & 67.45 & 64.40 & 86.65 & 61.58 \\
Qwen2.5-VL-7B (Perception+Reasoning) & 71.20 & 37.82 & 38.29 & 87.60 & 70.46 & 64.07 & 86.84 & 65.18 \\
Qwen2.5-VL-7B (Reasoning-only) & 70.00 & 36.55 & 36.86 & 86.64 & 69.80 & 62.80 & 84.97 & 63.95 \\
\hline
Qwen3-VL-8B & 72.40 & 31.09 & 50.86 & 87.86 & 70.85 & 70.00 & 88.11 & 67.31 \\
Qwen3-VL-8B (Perception+Reasoning) & 75.90 & 43.78 & 56.10 & 86.29 & 74.51 & 73.07 & 87.88 & 71.08 \\
Qwen3-VL-8B (Reasoning-only) & 73.80 & 42.26 & 56.10 & 86.90 & 73.86 & 71.67 & 87.35 & 70.28 \\
\hline
\end{tabular}
\label{tab:fig2_exact_numbers}
\end{table}

Table~\ref{tab:fig2_exact_numbers} reports the exact values corresponding to Figure~2. 
Across both Qwen2.5-VL-7B and Qwen3-VL-8B, incorporating perception data consistently yields larger gains on visual math benchmarks compared to reasoning-only training. 
For instance, on Qwen3-VL-8B, perception+reasoning improves MVerse (VI) from 42.26\% to 43.78\% and MVista from 73.80\% to 75.90\%, while achieving comparable performance on A-OKVQA and POPE. 
A similar trend is observed in Qwen2.5-VL-7B, where perception+reasoning outperforms reasoning-only on WeMath (38.29\% \textit{v.s.} 36.86\%) and MVerse (VI) (37.82\% \textit{v.s.} 36.55\%). 
These results indicate that strengthening visual perception directly translates into improved visual reasoning without degrading general perception capabilities.

\subsection{Visual Perception Data Example}
\label{sec:data_example}
\begin{figure}[t]
    \centering
    \includegraphics[width=1.0\linewidth]{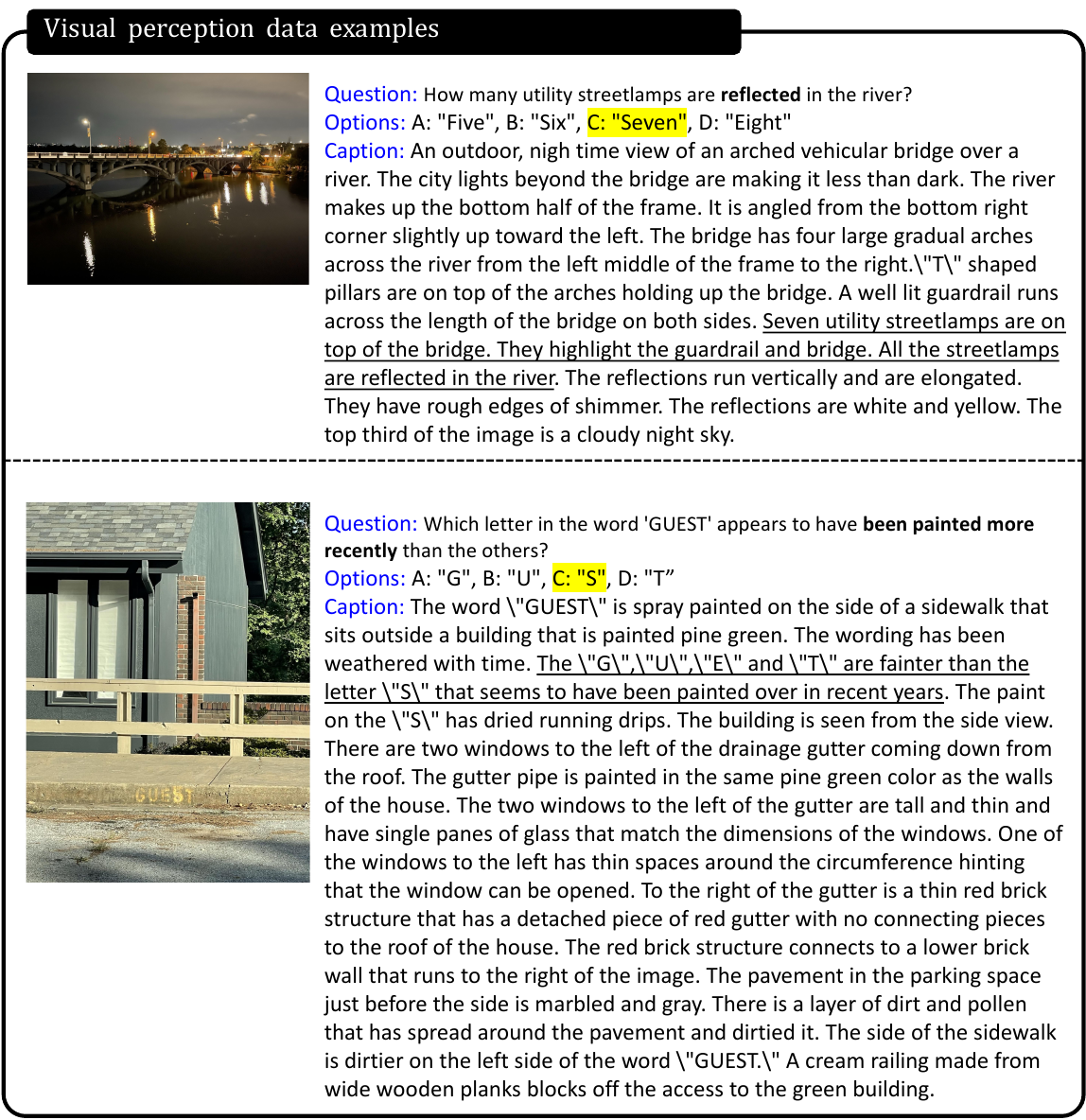}
    \caption{\textbf{Example of synthesized visual perception data.}}
    \label{fig:data_example}
\end{figure}

Here, we include two representative generated visual perception examples (Figure~\ref{fig:data_example}). The first requires robust \emph{object detection and counting under low-light conditions} by identifying seven streetlamps and their reflections on the river surface. The second targets \emph{fine-grained visual attribute discrimination}, where the model must infer the most recently painted letter in a weathered graffiti word based on color intensity and paint texture. 

Together, these examples illustrate that our generated perception data explicitly exercises core visual competencies such as object counting, reflection understanding, fine-grained appearance comparison, and material aging cues—capabilities that are often bottlenecks in downstream visual reasoning tasks.

\subsection{Prompt Settings}
\label{app:prompt_settings}
\begin{figure}[ht]
    \centering
    \includegraphics[width=1.0\linewidth]{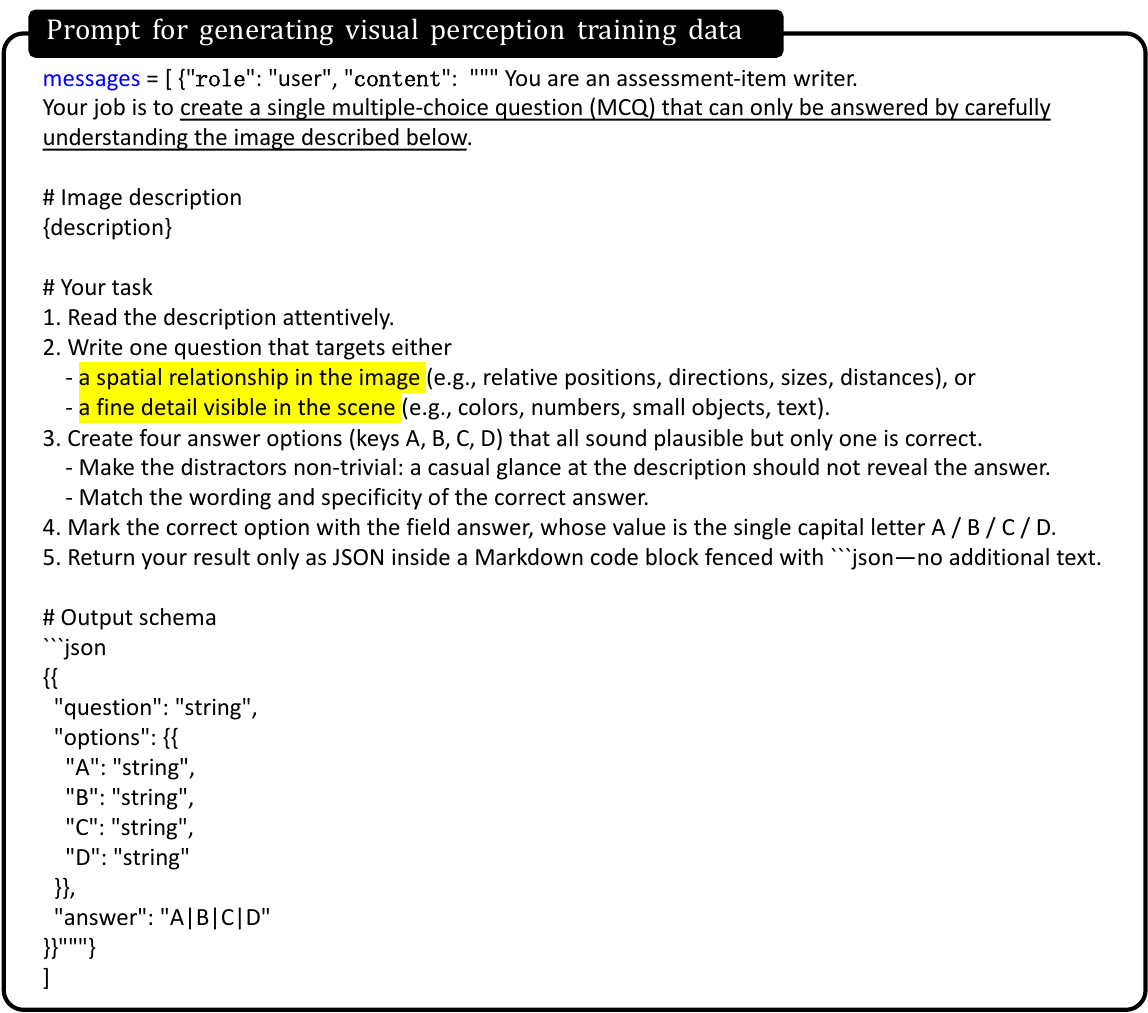}
    \caption{\textbf{Prompt for generating visual perception question-answer pairs.}}
    \label{fig:prompt_generation}
\end{figure}

\begin{figure}[ht]
    \centering
    \includegraphics[width=1.0\linewidth]{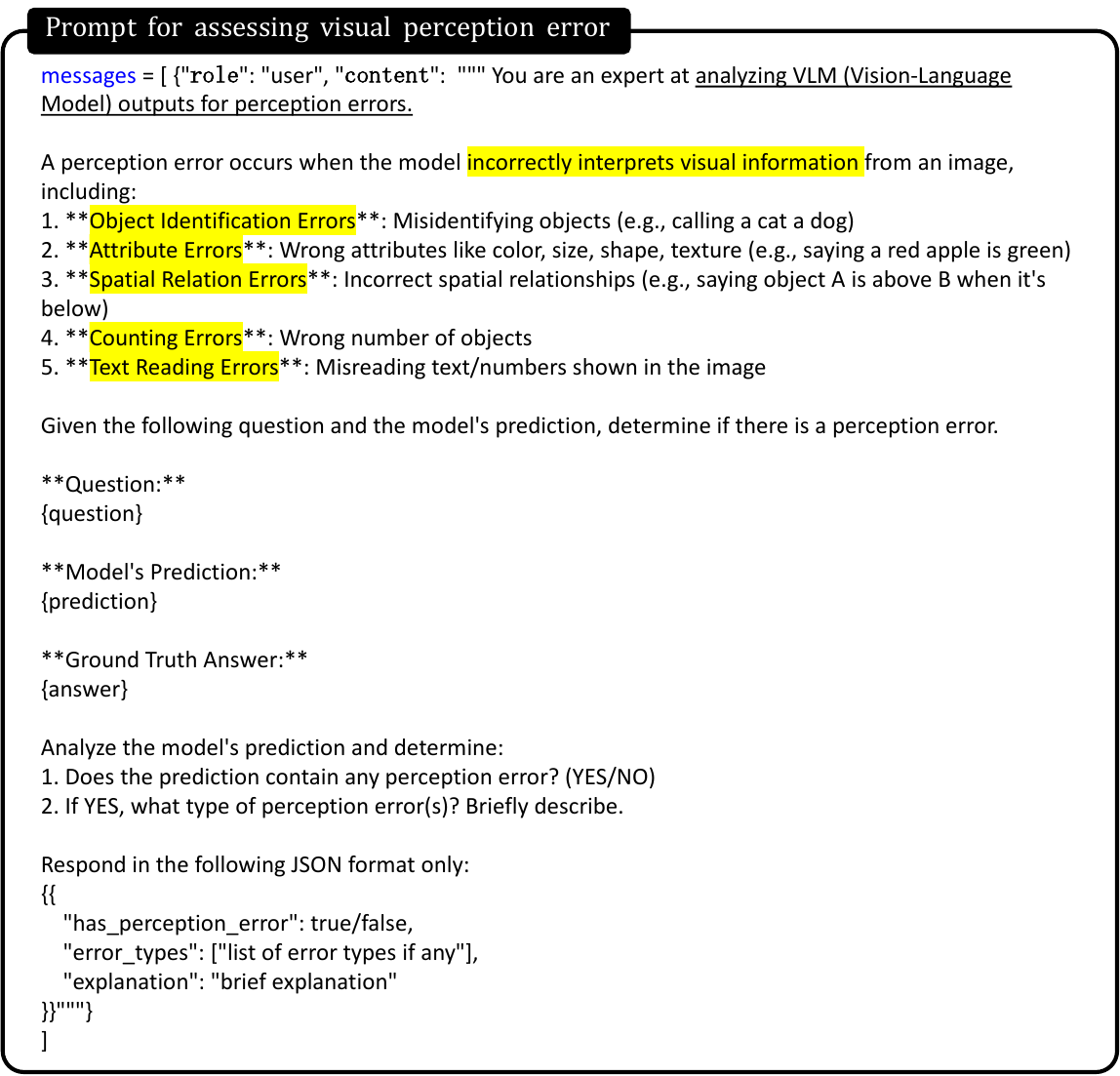}
    \caption{\textbf{Prompt for assessing visual perception errors in VLM's reasoning.}}
    \label{fig:prompt_assessing}
\end{figure}

\begin{figure}[ht]
    \centering
    \includegraphics[width=1.0\linewidth]{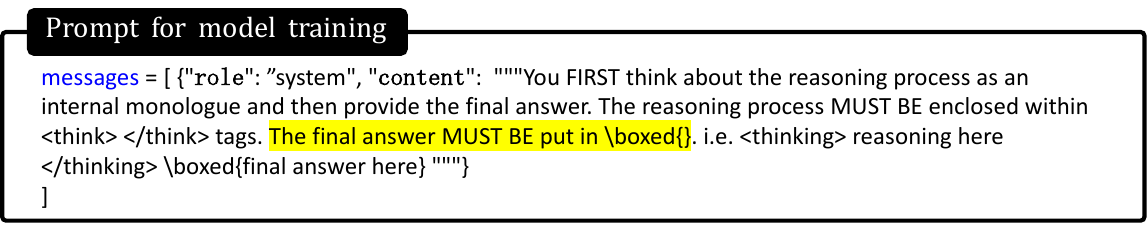}
    \caption{\textbf{System prompt used in our experiments.}}
    \label{fig:prompt_training}
\end{figure}

In this section, we provide all the prompts used in our experiments, including the prompt for (a) generating visual perception question-answering data (Figure~\ref{fig:prompt_generation}); (b) assessing visual perception errors in the model's reasoning (Figure~\ref{fig:prompt_assessing}); and (c) the system prompt used for model training (Figure~\ref{fig:prompt_training}).

\subsection{Extended Benchmark Results Across Four Model Families}
\label{app:extended_benchmarks}

\begin{table}[ht]
\centering
\small
\setlength{\tabcolsep}{1.8pt}
\caption{\textbf{Comprehensive comparison of base, merged, and staged training across four model families on extended benchmarks (Accuracy \%).} Best results within each model family are highlighted in \textbf{bold}.}
\resizebox{\columnwidth}{!}{%
\begin{tabular}{llccccccccccc}
\hline
\textbf{Base Model} & \textbf{Training} & \textbf{MVista} & \textbf{MVerse (VO)} & \textbf{MVerse (VI)} & \textbf{WeMath} & \textbf{DynaMath} & \textbf{MMStar} & \textbf{Hallusion} & \textbf{BLINK} & \textbf{VisOnlyQA} & \textbf{VStarBench} & \textbf{AVG} \\
\hline
\multirow{3}{*}{InternVL3.5-8B}
& Base & 60.70 & 11.93 & 5.08 & 40.10 & 44.85 & 43.27 & 39.88 & 46.66 & 36.89 & 43.98 & 37.33 \\
& Merged & 69.40 & 32.11 & 32.61 & 48.38 & \textbf{62.34} & \textbf{64.40} & 48.55 & 56.13 & 50.89 & \textbf{62.83} & 52.76 \\
& Staged & \textbf{70.30} & \textbf{33.76} & \textbf{34.26} & \textbf{49.90} & 62.83 & 61.60 & \textbf{53.47} & \textbf{57.71} & \textbf{52.00} & 61.26 & \textbf{53.71} \\
\hline
\multirow{3}{*}{InternVL3-8B}
& Base & 18.10 & 17.13 & 22.34 & 3.14 & 13.15 & 46.33 & 30.06 & \textbf{48.55} & 41.56 & 56.54 & 29.69 \\
& Merged & 60.70 & 25.25 & 30.46 & 25.05 & 45.79 & 51.87 & 30.35 & 48.45 & 40.22 & \textbf{61.26} & 41.94 \\
& Staged & \textbf{65.40} & \textbf{29.70} & \textbf{32.99} & \textbf{34.95} & \textbf{52.42} & \textbf{53.00} & \textbf{36.42} & 46.87 & \textbf{46.22} & 59.16 & \textbf{45.71} \\
\hline
\multirow{3}{*}{Qwen2.5-VL-7B}
& Base & 68.40 & 24.11 & 25.00 & 30.86 & 51.54 & 63.67 & \textbf{40.46} & \textbf{55.39} & 49.56 & \textbf{76.96} & 48.60 \\
& Merged & 69.75 & 29.57 & 34.23 & \textbf{37.24} & 52.89 & 63.33 & 36.56 & 54.60 & 47.95 & 76.83 & 50.30 \\
& Staged & \textbf{71.45} & \textbf{32.93} & \textbf{38.13} & 36.88 & \textbf{54.03} & \textbf{64.79} & 39.95 & 55.71 & \textbf{48.67} & 76.70 & \textbf{51.92} \\
\hline
\multirow{3}{*}{Qwen3-VL-8B}
& Base & 72.40 & 26.90 & 31.09 & 50.86 & \textbf{66.43} & 70.00 & 32.95 & \textbf{68.39} & 61.78 & \textbf{83.77} & 56.46 \\
& Merged & 73.80 & 36.68 & 40.36 & 55.43 & 54.15 & 70.60 & 53.18 & 61.34 & 61.56 & 80.63 & 58.77 \\
& Staged & \textbf{75.90} & \textbf{39.97} & \textbf{43.78} & \textbf{56.10} & 61.14 & \textbf{73.07} & \textbf{59.54} & 64.12 & \textbf{64.00} & \textbf{83.77} & \textbf{62.14} \\
\hline
\end{tabular}
}
\label{tab:extended_benchmarks}
\end{table}

Table~\ref{tab:extended_benchmarks} presents a comprehensive evaluation across four model families on ten extended benchmarks, including DynaMath~\cite{zou2024dynamath}, HallusionBench~\cite{guan2024hallusionbench}, BLINK~\cite{fu2024blink}, VisOnlyQA~\cite{kamoi2024visonlyqa}, V*Bench~\cite{wu2024vstarbench}, and CV-Bench~\cite{tong2024cvbench}. Staged training consistently outperforms merged training across all architectures: InternVL3-8B shows the largest gain (+3.77\% overall), followed by Qwen3-VL-8B (+3.37\%), Qwen2.5-VL-7B (+1.62\%), and InternVL3.5-8B (+0.95\%). Notably, for InternVL3-8B, staged training improves WeMath from 25.05\% to 34.95\% (+9.90\%), demonstrating that the benefit of decoupling perception and reasoning is especially impactful for weaker base models. These results confirm that our staged training paradigm generalizes beyond the Qwen family to architecturally distinct VLMs.

\subsection{Statistical Robustness: Three-Run Averaged Results}
\label{app:3run_results}

\begin{table}[ht]
\centering
\small
\setlength{\tabcolsep}{2.0pt}
\caption{\textbf{Three-run averaged results for Qwen3-VL-8B and Qwen2.5-VL-7B (Accuracy \%).} Staged training consistently outperforms merged training across all averaged benchmarks, demonstrating statistical robustness. Best results in each row pair are in \textbf{bold}.}
\resizebox{\columnwidth}{!}{%
\begin{tabular}{lcccccccccccccccc}
\hline
\textbf{Model} & \textbf{MVista} & \textbf{MVision} & \textbf{MVerse (VO)} & \textbf{MVerse (VI)} & \textbf{WeMath} & \textbf{DynaMath} & \textbf{A-OKVQA} & \textbf{RWQA} & \textbf{MMStar} & \textbf{Hallusion} & \textbf{POPE} & \textbf{BLINK} & \textbf{CV-Bench} & \textbf{VisOnlyQA} & \textbf{VStarBench} & \textbf{AVG} \\
\hline
Qwen2.5-VL-7B (Staged) & \textbf{71.50} & \textbf{20.94} & \textbf{32.95} & \textbf{38.28} & 37.02 & \textbf{54.05} & \textbf{87.07} & 69.54 & \textbf{64.82} & \textbf{39.98} & \textbf{87.06} & \textbf{55.74} & \textbf{75.95} & \textbf{48.82} & \textbf{76.79} & \textbf{57.37} \\
Qwen2.5-VL-7B (Merged) & 69.63 & 19.52 & 29.40 & 34.01 & \textbf{37.21} & 52.84 & 85.85 & \textbf{70.15} & 63.31 & 36.42 & 84.67 & 54.57 & 74.72 & 47.85 & 76.62 & 55.78 \\
\hline
Qwen3-VL-8B (Staged) & \textbf{76.20} & \textbf{28.84} & \textbf{40.01} & \textbf{43.23} & \textbf{56.86} & \textbf{68.67} & \textbf{86.99} & \textbf{74.08} & \textbf{73.15} & \textbf{55.88} & \textbf{87.63} & \textbf{65.40} & \textbf{80.70} & \textbf{64.08} & \textbf{85.51} & \textbf{65.82} \\
Qwen3-VL-8B (Merged) & 72.93 & 26.86 & 35.91 & 38.88 & 52.98 & 66.58 & 85.42 & 73.77 & 70.20 & 50.97 & 87.18 & 62.97 & 78.31 & 62.15 & 80.28 & 63.03 \\
\hline
\end{tabular}
}
\label{tab:3run_results}
\end{table}

Table~\ref{tab:3run_results} reports results averaged over three independent evaluation runs. Staged training outperforms merged training on 14/15 benchmarks for Qwen3-VL-8B (+2.79\% overall AVG) and 12/15 benchmarks for Qwen2.5-VL-7B (+1.59\% overall AVG). The few benchmarks where merged training leads (e.g., WeMath and RWQA for Qwen2.5-VL-7B) show differences within 0.6\%, well within noise. These results confirm that the improvements from staged training are statistically robust and not artifacts of evaluation variance.

\subsection{Response Length on Test Sets}
\label{app:response_length}

\begin{table}[ht]
\centering
\small
\caption{\textbf{Average response length (tokens) on visual math test sets for Qwen3-VL-8B.} Staged training produces shorter responses across all benchmarks while achieving higher accuracy (Table~\ref{tab:staged_vs_merged}).}
\begin{tabular}{lcccc}
\hline
\textbf{Model} & \textbf{MathVista} & \textbf{MathVision} & \textbf{MathVerse (VO)} & \textbf{WeMath} \\
\hline
Staged & 1325.89 & 2930.41 & 1541.89 & 1745.69 \\
Merged & 1420.30 & 3163.41 & 1764.93 & 1906.07 \\
\hline
Reduction & $-$6.6\% & $-$7.4\% & $-$12.6\% & $-$8.4\% \\
\hline
\end{tabular}
\label{tab:response_length}
\end{table}

Table~\ref{tab:response_length} shows that staged training produces 6.6--12.6\% shorter responses across all visual math test benchmarks compared to merged training, consistent with the training-time observation in Figure~\ref{fig:length_comparison}. Combined with the higher accuracy achieved by staged training, this confirms that stronger perception reduces the need for excessive reasoning and repeated image re-checking.

\end{document}

%% file: sections/intro.tex
\begin{figure}[t]
    \centering
    \includegraphics[width=1.0\linewidth]{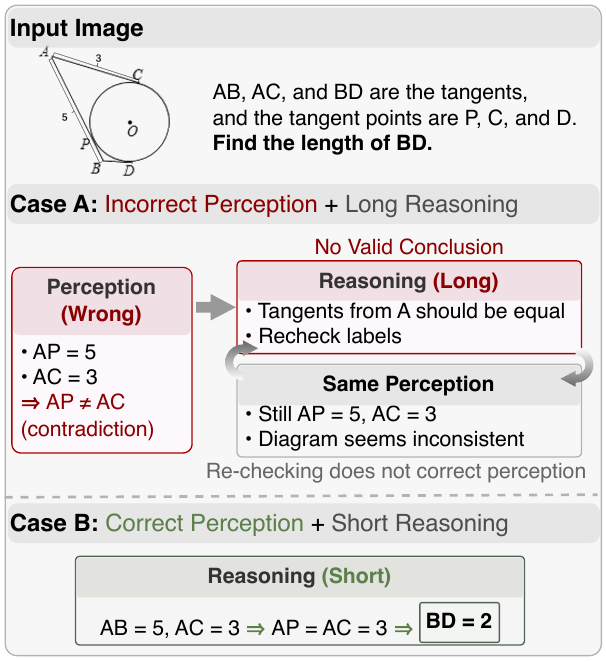}
    \caption{\textbf{Longer thinking can not fix incorrect perception.} Re-checking the image during the reasoning leads to the same perception error.}
    \label{fig:teaser}
\end{figure}

\section{Introduction}

Vision-Language Models (VLMs) have achieved remarkable progress in a wide range of multimodal tasks, including visual question answering~\cite{yue2024mmmu,huang2025medvlthinker,wu2025medreason}, diagram understanding~\cite{hou2024vision,hong2024cogvlm2}, and visual mathematical reasoning~\cite{liu2023visual,wang2024enhancing,xu2025llava}. Recent advances are largely driven by post-training techniques that emphasize long chain-of-thought reasoning via reinforcement learning (RL), enabling models to reason longer for better results~\cite{peng2025lmm,chen2025sft,zhan2025vision,shen2025vlm}. 

However, in many visual reasoning tasks, performance is not primarily limited by reasoning capability but by \emph{visual perception} --- \eg, visual mathematics~\cite{lindstrom2022clevr,zhuang2025math}, geometry problems~\cite{lu2023mathvista}, and diagram-based reasoning~\cite{mathew2021docvqa}.
We find that failures in VLM reasoning often stem from the very first visual perception step: once an error is introduced, subsequent reasoning rarely corrects it but instead compounds the mistake based on incorrect perceptual assumptions (see Case A in Figure~\ref{fig:teaser}).
In contrast, when visual perception is correct, the reasoning becomes concise and converges quickly to the correct answer (Case B). 
To validate this, we present an analysis of 3 visual math datasets by using the Claude-Haiku-4.5~\cite{anthropic2024claude3} to detect the perception errors in the VLM reasoning process: among all incorrectly sampled answers from Qwen3-VL-8B~\cite{bai2025qwen3vltechnicalreport}, \textbf{86.9\%} are due to the visual perception error as described.
Both qualitative and quantitative observations, complementing previous works~\cite{ogezi2025spare,zhu2026can,liu2025more}, highlight a key limitation of current post-training practices: \textit{longer reasoning does not compensate for incorrect perception}.

We hypothesize that the failure mode may result from flawed post-training paradigms, which emphasize visual reasoning training much more than visual perception in recent studies.
We argue that \textit{visual perception should be treated as an independent and fundamental capability in VLMs and trained separately}. 
To validate our hypothesis, we conduct comprehensive investigations by decoupling VLM capabilities into three  stages: visual perception, textual reasoning, and visual reasoning. We propose a staged post-training framework in which each capability is progressively refined using dedicated datasets. In the visual perception stage, we explore the transition from caption based supervised fine-tuning (SFT) to reinforcement learning with verifiable rewards (RLVR). To facilitate this, we construct a scalable data pipeline that transforms standard image-caption datasets~\cite{onoe2024docci} into structured, perception-focused training data, allowing the model to close the gap between raw visual input and textual alignment using fully open resources.

Our experimental findings highlight three key factors that are essential for effectively enhancing visual perception in VLMs: (a) \textbf{Dedicated data}, similar to textual and visual reasoning, visual perception is not a ``solved'' pre-training byproduct but requires further targeted optimization with specialized data. On the WeMath benchmark~\cite{qiao2025we}, incorporating the visual perception stage in post-training yields a 7.43-point accuracy gain over the Qwen2.5-VL-7B~\cite{bai2025qwen2} base model and also raises Qwen3-VL-8B performance from 50.9\% to 56.1\% (Section~\ref{sec:perception_data}); (b) \textbf{Staged training}: the staged training paradigm outperforms the common one-stage training setting in which all data for different capabilities are merged and shuffled during post-training. Our staged-trained Qwen3-VL-8B achieves a 1.46-point increase in math reasoning accuracy while producing 20.8\% shorter reasoning traces (Section~\ref{sec:staged_merged}) compared to the one-stage training. Moreover, the order of stage optimization is critical, as visual perception serves as the fundamental scaffold that should be solidified before refining visual reasoning. Disrupting this order reduces the average visual math performance of Qwen2.5-VL-7B from 42.3\% to 37.7\% (Section~\ref{sec:stage_order}); and (c) \textbf{RLVR-based visual perception learning}, RLVR provides a significantly more effective training signal for visual perception than caption-based SFT. While SFT can inadvertently degrade performance by imposing token-level, off-policy supervision from data that may be of lower quality than the pre-training corpus, RL keeps the model on-policy, resulting in better alignment. Substituting SFT for RL in visual perception training leads to drops of 8.1\% and 1.6\% in accuracy for the Qwen2.5-VL-7B and Qwen3-VL-8B models, respectively, on the WeMath benchmark (Section~\ref{sec:sft_vs_rl}).


Beyond these empirical findings, our work introduces a conceptual contribution: staged training by capability type can be viewed as \emph{capability-dimension curriculum learning}, a framework orthogonal to traditional difficulty-based curricula. We demonstrate that these two curriculum dimensions are complementary---combining capability-based staging with difficulty-based ordering yields a 4.43\% improvement over merged training, surpassing either dimension alone (Section~\ref{sec:capability_difficulty}).

Overall, our staged-training Qwen3-VL-8B attains strong performance on both visual math reasoning (75.9\% on MathVista and 56.1\% on WeMath) and visual perception (74.5\% on RealWorldQA) benchmarks (Table~\ref{tab:open_source_comparison}). Compared to OneThinker-8B, our model improves accuracy by 1.5\% on WeMath and 3.0\% on RealWorldQA. These findings indicate that integrating our visual perception data with staged-training paradigm yields more advanced reasoning capabilities in VLMs.

%% file: sections/related_works.tex
\section{Related Work}

\subsection{Reasoning Vision-Language Models}

Recent work increasingly targets visual reasoning in VLMs. A common SFT-based direction is to distill structured reasoning traces into the model~\cite{xu2024llavacot,zhang2024improvevisionlanguagemodel,thawakar2025llamavo1rethinkingstepbystepvisual,shao2024visualcotadvancingmultimodal,li2025visreasonlargescaledatasetvisual}. 
In parallel, as DeepSeek-R1~\citep{Guo_2025} gains success in textual reasoning by using Reinforcement Learning with Verifiable Rewards (RLVR)~\cite{shao2024deepseekmathpushinglimitsmathematical}, this paradigm has been adapted to multimodal reasoning to encourage exploration and self-correction~\cite{yang2025r1onevisionadvancinggeneralizedmultimodal,deng2025openvlthinkercomplexvisionlanguagereasoning,peng2025lmmr1empowering3blmms,feng2025videor1reinforcingvideoreasoning}. Typical vision-related tasks include general visual question answering (VQA)~\cite{marino2019okvqavisualquestionanswering,schwenk2022aokvqabenchmarkvisualquestion,hudson2019gqanewdatasetrealworld}, chart and infographic understanding~\cite{masry2022chartqabenchmarkquestionanswering,mathew2021infographicvqa}. 
Models trained on such tasks with RLVR are enabled to reason over multimodal inputs for higher accuracy. Our approach falls into the same category that leveraging the RLVR approach for tuning a competent reasoning VLM.




\begin{figure*}[t]
    \centering
    \includegraphics[width=1.0\linewidth]{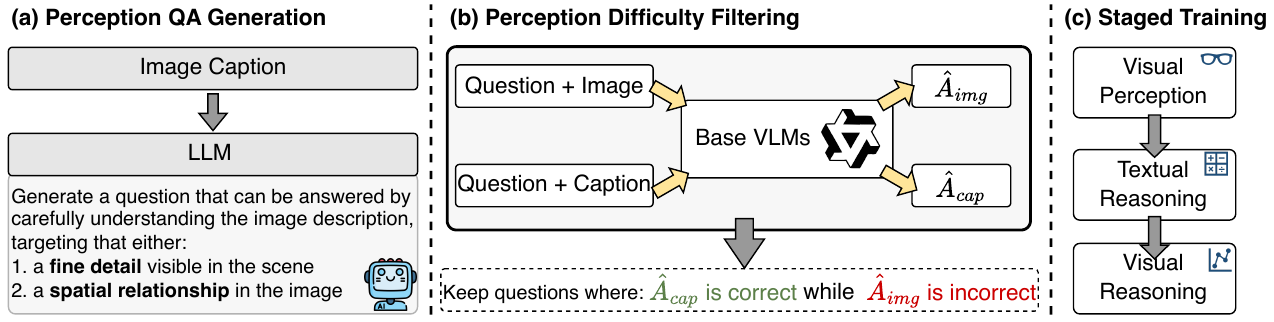}
    \caption{\textbf{Improving VLM Post-training with Visual Perception Data Synthesis and Staged Training}:
(a) Generating image-content based QA pairs by feeding captions to an LLM and labeling answers with a strong VLM;
(b) Perception difficulty filtering, which removes samples that can be answered by the base VLMs based on caption;
(c) Staged training by different capabilities \textit{from seeing to thinking}.
}
    \label{fig:pipeline}
\end{figure*}

\subsection{Post-training Paradigms For Reasoning VLMs}

Post-training for reasoning VLMs typically follows either \textbf{merged} training or \textbf{curriculum} training. 
In merged training, diverse supervision signals are merged and optimized together in a single phase. For SFT-based training, LLaVA-CoT exemplifies this by integrating multiple VQA sources with structured reasoning annotations in one training recipe~\cite{xu2024llavacot}. For RL-based training, VLAA-Thinker proposes Mixed Reward which blends grounding and reasoning rewards into a single-stage RL training~\citep{chen2025sft}.
Joint training is simple by design but lacks finer-grained considerations on the order of training data.
Curriculum learning fills the gap by training models on data with increasing difficulty, manifesting its effectiveness in works like Curr-ReFT~\cite{deng2025curr} and PC-GRPO~\cite{jeddi2025puzzle}, which boost performance on both reasoning and perception tasks.
Complementary to these training paradigms, recent diagnostic studies have specifically identified visual perception as a key bottleneck. VisOnlyQA~\cite{kamoi2024visonlyqa} reveals that models struggle with basic geometric understanding through vision-only questions, and NoReGeo~\cite{abdullaeva2026noregeo} isolates perception failures from reasoning by constructing non-reasoning geometry benchmarks. While these works focus on diagnosis, our work addresses the identified gap through a training methodology: instead of sorting data by difficulty, we propose a capability-based curriculum that decouples perception from reasoning and finds that capabilities should be learned following certain orders.

%% file: sections/method.tex
\section{Staged Post-training Pipeline}



\subsection{Data Synthesis and Curation} 
We construct three disjoint datasets corresponding to visual perception, textual reasoning, and visual reasoning, respectively. All datasets are synthesized or curated from fully open-source resources.
\subsubsection{Perception Data Synthesis}
The objective of the visual perception stage is to improve a model’s ability to
accurately recognize fine-grained visual details and relative spatial relations 
without requiring multi-step reasoning.

\vspace{2pt}\noindent\textbf{Question-Answer Generation from Captions.} 
We firstly collect image-caption pairs from the DOCCI dataset~\cite{onoe2024docci}, which contain 15K images paired with fine-grained captions.
As shown in Figure~\ref{fig:pipeline}(a), for each image-caption pair $(I, C)$, we prompt an LLM $f_{\text{gen}}$ (in this work, \texttt{Qwen2.5-72B}) to generate a set of perception-focused question-answer pairs:

\begin{equation}
    (Q,A) = f_\text{gen}(C)
\end{equation}

where each question $Q$ emphasizes visual details or spatial relations
that are explicitly grounded in the image.
The generated answer $A$ serves as the ground truth. The prompt we used is provided in Appendix Figure~\ref{fig:prompt_generation}.

\paragraph{Perception Difficulty Filtering.}
To isolate samples that specifically reflect perception deficiencies,
we introduce a perception-sensitive filtering criterion as illustrated in Figure~\ref{fig:pipeline}(b).
Let $f_{\theta}$ denote the base VLM.
For each generated question $Q$, we evaluate two inference pathways:
\begin{equation}
\hat{A}_{\text{img}} = f_{\theta}(I, Q), \quad
\hat{A}_{\text{cap}}= f_{\theta}(C, Q).
\end{equation}
Where $\hat{A}_\text{img}$ refers to the answer to $Q$ by $f_\theta$, with only image $I$ provided, and $\hat{A}_\text{cap}$ is the answer generated based on the paired caption.
We retain a sample $(I, Q, A)$ if and only if:
\begin{equation}
\mathbb{I}[\hat{A}_{\text{img}} \neq A] \land
\mathbb{I}[\hat{A}_{\text{cap}} = A],
\end{equation}
where $\mathbb{I}[\cdot]$ is the indicator function.
This condition ensures that the information required to answer $Q$
is present in the caption $C$, while the model fails when relying on its
own visual perception from $I$.

To further improve robustness, we apply this filtering using two models,
$f_{\theta}^{(1)} = \texttt{Qwen2.5-VL-7B}$ and
$f_{\theta}^{(2)} = \texttt{Qwen2.5-VL-32B}$.
The resulting dataset $\mathcal{D}_{\text{perc}}$ contains samples that are
challenging due to insufficient visual perception rather than reasoning ability. Detailed visual perception data examples are provided in Appendix~\ref{sec:data_example}.

\subsubsection{Reasoning Data Curation}
For textual reasoning, we use the open-source
\texttt{ORZ-Math-13k} dataset~\cite{hu2025open},
which consists of challenging math reasoning problems that require multi-step
logical inference without visual inputs.
The resulting textual reasoning dataset is denoted as $\mathcal{D}_{\text{text}}$.

For visual reasoning, we follow prior work in constructing challenging
multimodal reasoning datasets~\cite{chen2025sft,xu2025llava}.
We collect samples from multiple open-source sources, including
CLEVR-Math~\cite{lindstrom2022clevr}, GeoQA170K~\cite{gao2023g}, Math PUMA~\cite{zhuang2025math}, DocVQA~\cite{mathew2021docvqa}, and ArxivQA~\cite{li2024multimodal}.
We retain samples that require both accurate perception and multi-step reasoning,
forming the dataset $\mathcal{D}_{\text{vis}}$.

\subsection{Training Strategies}
\subsubsection{Staged Training}

We adopt Group Relative Policy Optimization (GRPO)~\cite{shao2024deepseekmathpushinglimitsmathematical} to enhance the model’s reasoning ability without relying on a separate value model. For each input $x$, a group of $G$ responses $\{y_i\}_{i=1}^{G}$ is sampled from the old policy $\pi_{\theta_{\text{old}}}$, and each response is assigned a composite reward $R(x,y_i)=r_{\text{acc}}(x,y_i)+r_{\text{format}}(x,y_i)$. 
The group-relative advantage is computed by standardizing rewards within each group as:
\begin{equation}
A_i = \frac{R(x,y_i)-\mu_R}{\sigma_R+\epsilon},
\end{equation}
where $\mu_R$ and $\sigma_R$ denote the group mean and standard deviation. The policy is then optimized to maximum clipped  objective with KL regularization:
\begin{equation}
\begin{aligned}
\mathcal{J}_{\mathrm{GRPO}}&(\theta)
= \\ & \mathbb{E}_{x,y}\! 
\left[ \frac{1}{G}
\sum_{i=1}^{G}
\min\!\big(\rho_i A_i,\;\mathrm{clip}(\rho_i,1-\epsilon,1+\epsilon)A_i\big)
\right] 
\\
&- \beta\,\mathrm{KL}(\pi_\theta\|\pi_{\text{ref}}),
\end{aligned}
\label{eq:grpo_obj}
\end{equation}
where $\rho_i=\pi_\theta(y_i|x)/\pi_{\theta_{\text{old}}}(y_i|x)$ and $\pi_{\text{ref}}$ is the reference policy from supervised fine-tuning.


In staged training, we optimize the model sequentially over three stages. Each stage is trained for the same number of epochs using identical hyperparameters. 
The training order is denoted as:
\begin{equation}
\mathcal{D}_{\text{perc}} \rightarrow \mathcal{D}_{\text{text}} \rightarrow\mathcal{D}_{\text{vis}}.
\end{equation}

\subsubsection{Merged Training}
\label{sec:merged_training}
For comparison, we construct a merged training baseline by combining
all datasets: $\mathcal{D}_{\text{merged}} =\mathcal{D}_{\text{perc}} \cup\mathcal{D}_{\text{text}} \cup\mathcal{D}_{\text{vis}}.$
The model is trained on $\mathcal{D}_{\text{merged}}$ with identical hyperparameters and the same total number of steps,
reflecting common post-training practices in which perception and reasoning supervision are jointly optimized.

%% file: sections/experiment.tex
\section{Experimental Analysis}

\subsection{Experimental Setup}
\paragraph{Models.}
We conduct experiments on two
VLM backbones Qwen3-VL-8B-Instruct~\cite{bai2025qwen3vltechnicalreport} and Qwen2.5-VL-7B-Instruct~\cite{bai2025qwen2}. 
In addition, we further benchmark our staged-training models
against a diverse set of open-weight reasoning VLMs.
Specifically, for models built upon Qwen2.5-VL-7B, we include GThinker~\cite{zhan2025gthinker}, MMR1~\cite{leng2025mmr1}, OpenVLThinker~\cite{deng2025openvlthinker}, R1-OneVision-RL~\cite{yang2025r1}, and WeThink~\cite{yang2025wethink} as baselines. 
For models based on Qwen3-VL-8B, we compare against the OneThinker~\cite{feng2025onethinker}. 
These baselines represent recent efforts that emphasize visual reasoning, reinforcement learning, or long-chain-of-thought generation, making them strong and relevant comparators for our study.
All baseline models are evaluated under their officially released configurations.

\paragraph{Hyperparameter Setting.}
We adopt EasyR1~\cite{zheng2025easyr1} as the training framework across all experiments. The system prompt used during training is fixed and provided in Appendix~\ref{app:prompt_settings}. The maximum response length is set to 2048 tokens, and the sampled group size in Equation~\ref{eq:grpo_obj} is fixed at 5. All experiments are conducted on a server with 8 NVIDIA H200 GPUs.

For staged training, visual encoder is enabled for all stages.
The number of training steps for the three stages is set to 90, 375, and 465, respectively, ensuring that each stage has the same number of training epochs. 
For the merged training baseline (Section~\ref{sec:merged_training}),
the visual encoder is disabled throughout training, following common practice in reasoning-focused post-training~\cite{chen2025sft,yang2025wethink}.
The merged training baseline is trained for 930 steps, matching the total number of training steps used in staged training. More details about the hyperparameter setting are provided in Section~\ref{app:hyperparameter}.

\begin{figure}[ht]
    \centering
    \includegraphics[width=1.0\linewidth]{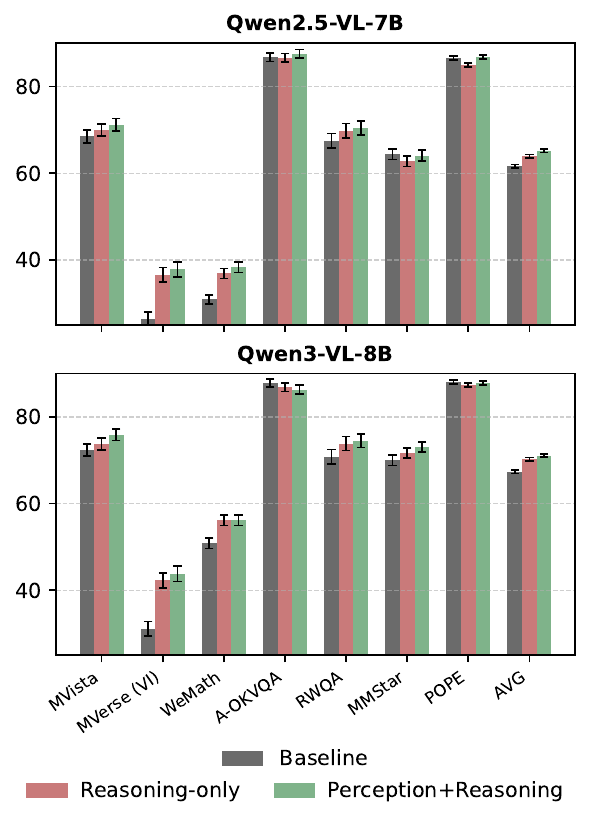}
    \caption{\textbf{Comparison between the base model, the model trained with reasoning-only, and perception+reasoning data.} Incorporating perception data improves visual math while maintaining perception capabilities. We show standard error bars here, and the exact values are provided in Appendix~\ref{app:fig2_exact_numbers}.
    \label{fig:perception_w_wo}
    }
\end{figure}

\paragraph{Benchmarks.}
We evaluate model performance on a comprehensive suite of vision-language benchmarks, covering both visual math reasoning and general visual perception as listed as follow:

\begin{itemize}
    \item For \textbf{visual math reasoning}, we consider MathVista MINI~\citep[MVista;][]{lu2023mathvista},
    MathVision MINI~\citep[MVision;][]{wang2024measuring}, MathVerse Vision Intensive subset~\citep[MVerse (VI);][]{zhang2024mathverse}, and WeMath~\cite{qiao2025we}. 
    \item For \textbf{perception-oriented}, we include A-OKVQA~\cite{schwenk2022okvqa}, RealWorldQA (RWQA)~\cite{xAI2024RealworldQA}, MMStar~\cite{chen2024we}, and POPE~\cite{li2023evaluating}, which assess object recognition, commonsense understanding, real-world perception, and robustness to visual hallucination.
\end{itemize}
All evaluations are conducted using VLMEvalKit~\cite{duan2024vlmevalkit} as the unified evaluation codebase. We employ Claude-Haiku-4.5~\cite{anthropic2024claude3} as the judge model for all evaluated models and benchmarks.


\begin{table*}[t]
\centering
\small
\setlength{\tabcolsep}{2.8pt}
\caption{
\textbf{Comparison with representative open-weight VLMs (Accuracy \%).}
Accuracies (\%) are
reported on individual benchmarks as well as average scores.
Best results in each column are highlighted in \textbf{bold}, and second-best results are \underline{underlined}.
}
\begin{tabular}{l c c c c c c c c c c c}
\toprule
\textbf{Model} &
\multicolumn{5}{c}{\textbf{Visual Math}} &
\multicolumn{5}{c}{\textbf{Perception}} &
\textbf{Overall} \\
\cmidrule(lr){2-6} \cmidrule(lr){7-11} \cmidrule(lr){12-12}
& MVista & MVision & MVerse (VI) & WeMath & AVG
& A-OKVQA & RWQA & MMStar & POPE & AVG
& AVG \\
\midrule

Qwen2.5-VL-7B
& 68.50 & 22.37 & 26.40 & 30.86 & 37.03
& 86.81 & 67.45 & 64.40 & 86.65 & 76.33
& 56.68 \\

GThinker-7B
& 69.70 & 23.03 & 36.80 & 35.15 & 41.17
& 86.72 & 69.80 & 64.60 & \textbf{88.38} & 77.38
& 59.27 \\

MMR1-7B
& 67.40 & 22.04 & 27.03 & 44.67 & 40.29
& 85.07 & 64.31 & 63.67 & 85.78 & 74.71
& 57.50 \\

OpenVLThinker-7B
& 70.60 & 22.70 & 36.29 & 35.90 & 41.37
& \textbf{88.73} & 69.41 & 63.40 & 80.82 & 75.59
& 58.48 \\

R1-OneVision-RL-7B
& 61.70 & 22.04 & 25.25 & 29.71 & 34.68
& 83.58 & 63.40 & 58.13 & 82.11 & 71.81
& 53.24 \\

WeThink-7B
& 69.50 & 23.03 & 34.39 & 46.57 & 43.37
& \underline{88.65} & 69.54 & 64.67 & 84.90 & 76.94
& 60.16 \\

Qwen2.5-VL-7B (Staged)
& 71.20 & 21.71 & 37.82 & 38.29 & 42.26
& 87.60 & 70.46 & 64.07 & 86.84 & 77.24
& 59.75 \\

\midrule
Qwen3-VL-8B
& 72.40 & 26.32 & 31.09 & 50.86 & \underline{45.17}
& 87.86 & 70.85 & 70.00 & \underline{88.11} & \underline{79.21}
& 62.19 \\

OneThinker-8B
& \underline{75.10} & \textbf{33.22} & \underline{41.50} & \underline{54.57} & \textbf{51.10}
& 86.72 & \underline{71.50} & \underline{70.20} & 86.14 & 78.64
& \underline{64.87} \\

Qwen3-VL-8B (Staged)
& \textbf{75.90} & \underline{28.62} & \textbf{43.78} & \textbf{56.10} & \textbf{51.10}
& 86.29 & \textbf{74.51} & \textbf{73.07} & 87.88 & \textbf{80.44}
& \textbf{65.77} \\

\bottomrule
\end{tabular}
\label{tab:open_source_comparison}
\end{table*}

\begin{table*}[t]
\centering
\small
\setlength{\tabcolsep}{2.7pt}
\caption{
\textbf{Comparison of merged and staged training on the same base VLM across visual math and perception benchmarks (Accuracy \%).} Accuracies (\%) are
reported on individual benchmarks as well as average scores. 
Best results in each column are highlighted in \textbf{bold}, and second-best results are \underline{underlined}.
}
\begin{tabular}{l l c c c c c c c c c c c}
\toprule
\textbf{Base Model} & \textbf{Training} &
\multicolumn{5}{c}{\textbf{Visual Math}} &
\multicolumn{5}{c}{\textbf{Perception}} &
\textbf{Overall} \\
\cmidrule(lr){3-7} \cmidrule(lr){8-12} \cmidrule(lr){13-13}
& & MVista & MVision & MVerse (VI) & WeMath & AVG
& A-OKVQA & RWQA & MMStar & POPE & AVG
& AVG \\
\midrule

\multirow{3}{*}{Qwen2.5-VL-7B}
& Base
& 68.50 & \textbf{22.37} & 26.40 & 30.86 & 37.03
& 86.81 & 67.45 & \textbf{64.40} & 86.65 & 76.33
& 56.68 \\

& Merged
& 70.00 & 20.39 & 35.15 & 37.43 & 40.74
& 86.03 & 69.28 & 63.73 & 84.74 & 75.95
& 58.34 \\

& Staged
& \textbf{71.20} & 21.71 & \textbf{37.82} & \textbf{38.29} & \textbf{42.26}
& \textbf{87.60} & \textbf{70.46} & 64.07 & \textbf{86.84} & \textbf{77.24}
& \textbf{59.75} \\

\midrule
\multirow{3}{*}{Qwen3-VL-8B}
& Base
& 72.40 & 26.32 & 31.09 & 50.86 & 45.17
& \textbf{87.86} & 70.85 & 70.00 & \textbf{88.11} & 79.21
& 62.19 \\

& Merged
& 73.80 & \textbf{28.95} & 40.36 & 55.43 & 49.64
& 85.50 & \textbf{75.56} & 70.60 & 87.19 & 79.71
& 64.67 \\

& Staged
& \textbf{75.90} & 28.62 & \textbf{43.78} & \textbf{56.10} & \textbf{51.10}
& 86.29 & 74.51 & \textbf{73.07} & 87.88 & \textbf{80.44}
& \textbf{65.77} \\

\bottomrule
\end{tabular}
\label{tab:staged_vs_merged}
\end{table*}

\subsection{The Vital Role of Visual Perception in Staged Post-training.}
\label{sec:perception_data}

To validate the necessity of visual-dedicated data, we employ a staged, decoupled training pipeline that first establishes a perceptual foundation before introducing complex reasoning. We evaluate this approach through two lenses: an internal ablation on data composition and a broad comparison with strong open-weight baselines.

\paragraph{The Impact of Visual Perception Data within Staged Training.} 
We first investigate whether reasoning data alone is sufficient during the post-training stages. We compare three configurations across Qwen2.5-VL-7B and Qwen3-VL-8B: the base models, a reasoning-only staged version (textual and visual), and our proposed incorporation of perception and reasoning data (Figure~\ref{fig:perception_w_wo}). 
Across both backbones, the reasoning-only post-training significantly enhances visual math performance; for Qwen2.5-VL-7B, MVerse (VI) and WeMath improve by 10.2\% and 6.0\%, respectively. 
However, excluding perception data introduces a ``perceptual tax''~\cite{liu2025more}. On Qwen2.5-VL-7B, reasoning-only training actually reduces MMStar performance by 1.6\%.In contrast, incorporating our visual perception data restores and exceeds base model integrity. By including perception tasks in the staged pipeline, RWQA scores climb to 70.5\% (+3.0\%) on Qwen2.5-VL-7B and 74.5\% (+3.6\%) on Qwen3-VL-8B. These results confirm that visual perception data is a fundamental prerequisite for balancing reasoning gains without sacrificing the model's eyes.

\paragraph{Performance Superiority of Perception-First Training.} 
To demonstrate the robustness of this decoupled pipeline, we compare our ``visual-perception-first'' models against specialized open-weight VLMs in Table~\ref{tab:open_source_comparison}. By prioritizing a solid perceptual foundation before scaling reasoning complexity, we achieve superior results without the trade-offs seen in existing models.
In the 7B category, our approach achieves a visual math average of 42.3\%, outperforming specialized reasoning baselines like GThinker, OpenVLThinker, and MMR1. Crucially, it maintains a superior average perception score of 77.2\%, proving that reasoning capabilities can be scaled more robustly when decoupled from perception.

The advantages are even more pronounced in the Qwen3-VL-8B series. Our staged-training model establishes new state-of-the-art benchmarks for 8B-parameter VLMs, leading in WeMath (56.1\%), MathVista (75.9\%), MMStar (73.1\%), and RealWorldQA (74.5\%). These improvements culminate in a record overall average of 65.8\%, surpassing both the base model and the reasoning-specialized baseline, OneThinker-8B. 
These results highlight that explicitly prioritizing visual perception in a staged pipeline is the key to scaling high-performance, general-purpose VLMs.



\subsection{Beyond One-stage Training: Analyzing Staged Training Paradigms and Ordering}
\label{sec:staged_training}

Our training paradigm decomposes VLM post-training into three distinct stages, each targeting a specific capability: visual perception (\textbf{Stage 1}), textual reasoning (\textbf{Stage 2}), and visual reasoning (\textbf{Stage 3}).
In this section, we conduct a thorough analysis of this staged training strategy. We begin by comparing it to the conventional single-stage paradigm, where data for all capabilities are combined into one dataset and optimized jointly (\textbf{merged training}) as depicted in Section~\ref{sec:merged_training}. 

We show that staged training not only delivers higher overall performance but also improves the optimization of visual perception, thereby reducing the cost of reasoning (see Section~\ref{sec:staged_merged}). In addition, we find that the advantage of staged training depends on the order of the stages: visual perception should be regarded as a more fundamental ability and optimized prior to visual reasoning (see Section~\ref{sec:stage_order}).

\begin{figure*}[t]
    \centering
    \includegraphics[width=0.95\linewidth]{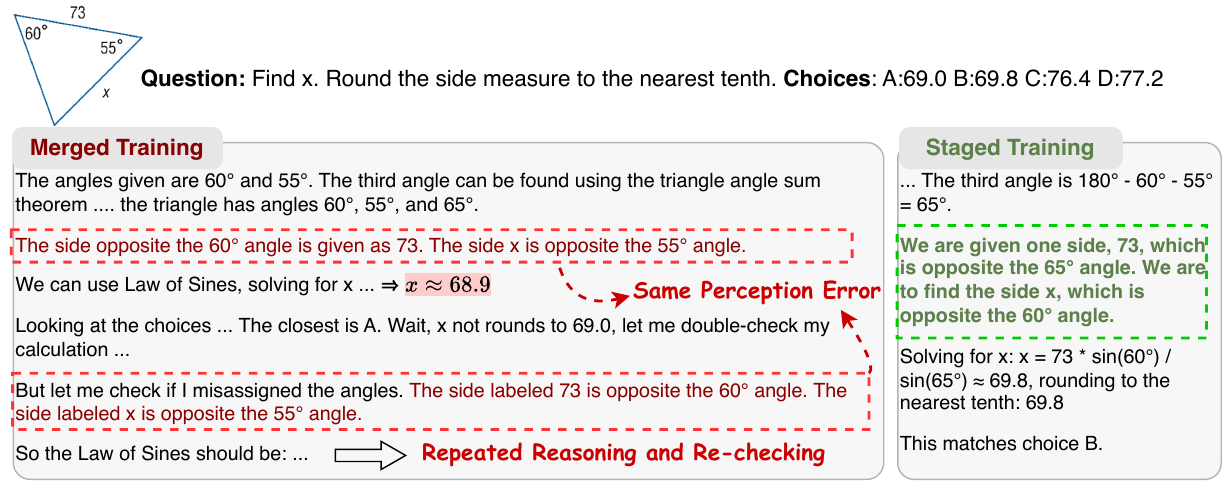}
    \caption{\textbf{Case Study between Staged and Merged Training Models.} The staged training model generates concise reasoning with correct perception.}
    \label{fig:case_study}
\end{figure*}

\begin{figure}[t]
    \centering
    \includegraphics[width=1.0\linewidth]{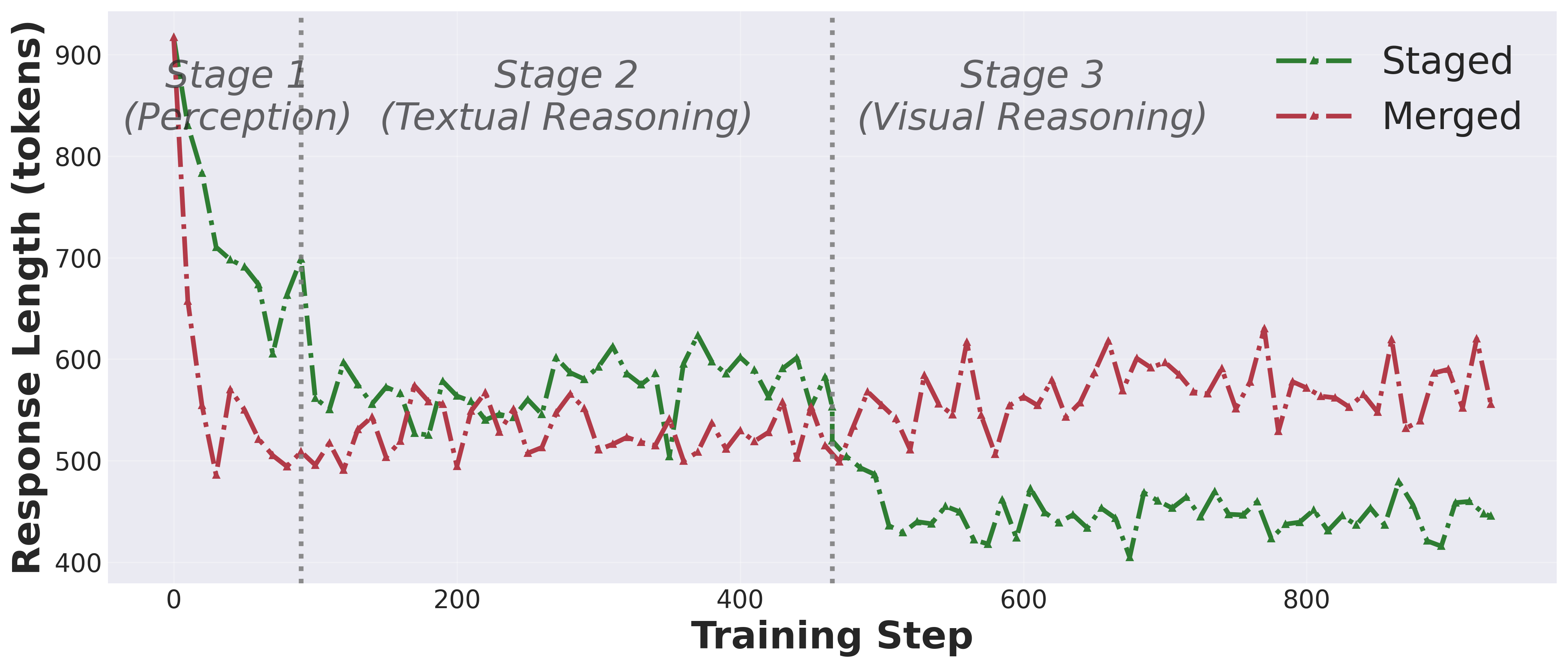}
    \caption{\textbf{Staged Training Reduces the Response Length for Visual Reasoning.} For the Qwen3-VL-8B model, we plot the average response length on the validation set over training steps, comparing the staged and merged training strategies.}
    \label{fig:length_comparison}
\end{figure}
\vspace{-2pt}

\subsubsection{Staged versus Merged Training}
\label{sec:staged_merged}

\paragraph{Overall Performance Comparison.}
We compare the base models, models with merged training, and those with staged training across visual math perception benchmarks (Table~\ref{tab:staged_vs_merged}).
Across both base models, staged training consistently achieves the best overall performance, demonstrating its general effectiveness.
For Qwen2.5-VL-7B, staged training improves the average visual math score from 37.0\% (base) and 40.7\% (merged) to 42.3\%, with clear gains on MVerse (26.4\% $\rightarrow$ 37.9\%) and WeMath (30.9\% $\rightarrow$ 38.3\%). Perception performance is also improved, increasing the average score to 77.2\%, compared to 76.3\% (base) and 76.0\% (merged), resulting in the highest overall score of 59.8\%.

Similar trends are observed for models with the Qwen3-VL-8B backbone. Staged training outperforms both base and merged training on visual math, improving the average score from 45.2\% to 51.1\% and achieving the best perception (average 80.4\%). Consequently, staged training attains the highest overall score (65.8\%) among all variants.
To further verify the generality of staged training beyond the Qwen family, we evaluate on InternVL3.5-8B and InternVL3-8B (Appendix~\ref{app:extended_benchmarks}). Staged training consistently outperforms merged training across both InternVL architectures, with overall gains of +0.95\% and +3.77\%, respectively, confirming that the benefit of decoupling perception and reasoning generalizes across different VLM backbones. We further validate statistical robustness by averaging over three independent runs across 15 benchmarks (Appendix~\ref{app:3run_results}); staged training wins on 14/15 benchmarks for Qwen3-VL-8B.

In addition, we employ Claude-4.5-Haiku to assess perception errors in the model’s reasoning, with the complete prompt provided in the Appendix~\ref{app:prompt_settings}.
We randomly selected 40 judgments from Claude-4.5-Haiku and manually verified whether it correctly identifies visual perception errors. The validation shows that 33/40 (82.5\%) of the samples match human judgments, suggesting that the Claude model is reliable for detecting visual perception errors.
For the Qwen3-VL-8B model, 857 out of 3044 samples from the MVista, MVision, and WeMath benchmarks are identified as having perception errors. After merged training, this count drops to 805, and staged training further decreases it to 781, indicating that explicitly decoupling perception and reasoning during training leads to more effective and robust VLM performance.  

\paragraph{Staged Training Leads to Better Perception and Shorter Thinking Costs.}
Figure~\ref{fig:length_comparison} shows the average response length during training for Qwen3-VL-8B under staged and merged training. While both approaches start with long responses, the staged model gradually reduces its response length as perception training progresses. During Stage~2, staged training maintains response lengths comparable to merged training, indicating that shorter outputs are not caused by suppressed reasoning.
A clear divergence appears in Stage~3, where the staged model produces responses that are 20.8\% shorter than those from merged training (average length 445 tokens \textit{v.s.} 562 tokens across the validation set), while achieving higher math reasoning accuracy, as shown in Table~\ref{tab:staged_vs_merged} (51.1\% \textit{v.s.} 49.6\%). This reduction is consistent at test time: on four visual math benchmarks, staged training produces 6.6--12.6\% shorter responses (Appendix~\ref{app:response_length}).

Figure~\ref{fig:case_study} provides a detailed comparison between merged and staged training. Under merged training, the model incorrectly assigns the side of length 73 to the wrong angle. This perceptual error persists across repeated image checks, leading to long and repeated reasoning traces without resolving the inconsistency.
In contrast, the staged-trained model correctly identifies the geometric relationships at the outset. With accurate perception, the subsequent reasoning becomes concise and directly yields the correct answer, explaining the shorter response lengths observed in Figure~\ref{fig:length_comparison}.

\begin{table*}[t]
\centering
\small
\setlength{\tabcolsep}{2.6pt}
\caption{
\textbf{Effect of stage order on visual math and perception performance (Accuracy \%).}
Best results in each column are highlighted in \textbf{bold}.
We compare different stage orders for staged training on Qwen2.5-VL-7B and Qwen3-VL-8B.
Stage~1$\rightarrow$2$\rightarrow$3 (perception $\rightarrow$ textual reasoning $\rightarrow$ visual reasoning)
and Stage~2$\rightarrow$1$\rightarrow$3 achieve comparable and consistently strong performance,
while reversing the order to Stage~3$\rightarrow$2$\rightarrow$1 leads to clear degradation
in both visual math and perception metrics.
}
\begin{tabular}{l c c c c c c c c c c c}
\toprule
\textbf{Model / Stage Order} &
\multicolumn{5}{c}{\textbf{Visual Math}} &
\multicolumn{5}{c}{\textbf{Perception}} &
\textbf{Overall} \\
\cmidrule(lr){2-6} \cmidrule(lr){7-11} \cmidrule(lr){12-12}
& MVista & MVision & MVerse (VI) & WeMath & AVG
& A-OKVQA & RWQA & MMStar & POPE & AVG
& AVG \\
\midrule

Qwen2.5-VL-7B (Base)
& 68.50 & 22.37 & 26.40 & 30.86 & 37.03
& 86.81 & 67.45 & 64.40 & 86.65 & 76.33
& 56.68 \\

Qwen2.5-VL-7B ( 1$\rightarrow$2$\rightarrow$3)
& \textbf{71.20} & 21.71 & \textbf{37.82} & \textbf{38.29} & 42.26
& \textbf{87.60} & \textbf{70.46} & 64.07 & \textbf{86.84} & \textbf{77.24}
& \textbf{59.75} \\

Qwen2.5-VL-7B ( 2$\rightarrow$1$\rightarrow$3)
& 71.70 & 23.36 & 38.20 & 38.38 & \textbf{42.91}
& 86.46 & 69.54 & \textbf{64.67} & 84.54 & 76.30
& 59.61 \\

Qwen2.5-VL-7B ( 3$\rightarrow$2$\rightarrow$1)
& 66.60 & \textbf{24.34} & 32.99 & 26.86 & 37.70
& 79.48 & 68.63 & 62.87 & 85.70 & 74.17
& 55.93 \\

\midrule

Qwen3-VL-8B (Base)
& 72.40 & 26.32 & 31.09 & 50.86 & 45.17
& 87.86 & 70.85 & 70.00 & \textbf{88.11} & 79.21
& 62.19 \\

Qwen3-VL-8B ( 1$\rightarrow$2$\rightarrow$3)
& \textbf{75.90} & \textbf{28.62} & \textbf{43.78} & 56.10 & \textbf{51.10}
& 86.29 & 74.51 & 73.07 & 87.88 & 80.44
& 65.77 \\

Qwen3-VL-8B ( 2$\rightarrow$1$\rightarrow$3)
& 74.90 & \textbf{28.62} & 43.65 & 55.81 & 50.75
& 86.72 & \textbf{76.21} & \textbf{73.33} & 87.17 & \textbf{80.86}
& \textbf{65.80} \\

Qwen3-VL-8B ( 3$\rightarrow$2$\rightarrow$1)
& 75.20 & 26.64 & 40.86 & \textbf{57.43} & 50.03
& 84.45 & 74.77 & 71.33 & 87.65 & 79.55
& 64.79 \\

\bottomrule
\end{tabular}
\label{tab:stage_order}
\end{table*}

\begin{table*}[t]
\centering
\small
\setlength{\tabcolsep}{2.8pt}
\caption{
\textbf{Effect of reinforcement learning versus supervised fine-tuning for Stage~1 perception training (Accuracy \%).}
Best results in each column are highlighted in \textbf{bold}.
Across both Qwen2.5-VL-7B and Qwen3-VL-8B, RLVR consistently yields higher perception accuracy and leads to stronger downstream visual math performance, resulting in improved overall accuracy.
}
\begin{tabular}{l c c c c c c c c c c c}
\toprule
\textbf{Model / Stage 1 Method} &
\multicolumn{5}{c}{\textbf{Visual Math}} &
\multicolumn{5}{c}{\textbf{Perception}} &
\textbf{Overall} \\
\cmidrule(lr){2-6} \cmidrule(lr){7-11} \cmidrule(lr){12-12}
& MVista & MVision & MVerse (VI) & WeMath & AVG
& A-OKVQA & RWQA & MMStar & POPE & AVG
& AVG \\
\midrule


Qwen2.5-VL-7B (RLVR)
& \textbf{71.20} & \textbf{21.71} & \textbf{37.82} & \textbf{38.29} & \textbf{42.26}
& \textbf{87.60} & \textbf{70.46} & \textbf{64.07} & \textbf{86.84} & \textbf{77.24}
& \textbf{59.75} \\

Qwen2.5-VL-7B (SFT)
& 66.40 & 18.75 & 32.87 & 30.10 & 37.03
& 84.63 & 70.20 & 62.60 & 85.26 & 75.67
& 56.35 \\

\midrule

Qwen3-VL-8B (RLVR)
& \textbf{75.90} & 28.62 & \textbf{43.78} & \textbf{56.10} & \textbf{51.10}
& \textbf{86.29} & \textbf{74.51} & \textbf{73.07} & \textbf{87.88} & \textbf{80.44}
& \textbf{65.77} \\

Qwen3-VL-8B (SFT)
& 74.70 & \textbf{32.24} & 42.89 & 54.48 & 51.08
& 85.94 & 72.29 & 72.60 & 86.00 & 79.21
& 65.14 \\

\bottomrule
\end{tabular}
\label{tab:rl_vs_sft}
\end{table*}

\begin{table*}[t]
\centering
\small
\setlength{\tabcolsep}{2.5pt}
\caption{
\textbf{Effect of combining capability-based and difficulty-based curricula on Qwen3-VL-8B (Accuracy \%).}
Best results in each column are highlighted in \textbf{bold}.
}
\begin{tabular}{lccccccccc}
\toprule
\textbf{Curriculum} & \textbf{MVision} & \textbf{MVerse (VO)} & \textbf{WeMath} & \textbf{DynaMath} & \textbf{RWQA} & \textbf{CV-Bench} & \textbf{V*Bench} & \textbf{AVG} \\
\midrule
None (Merged) & 28.95 & 36.68 & 55.43 & 54.15 & 75.56 & 78.53 & 80.63 & 58.56 \\
Capability & 28.62 & 39.97 & 56.10 & 61.14 & 74.51 & 79.62 & 83.77 & 60.53 \\
Difficulty & 24.67 & 40.86 & 54.10 & 67.07 & 72.68 & 79.36 & 83.77 & 60.36 \\
Capability+Difficulty & \textbf{33.22} & \textbf{41.75} & \textbf{57.43} & \textbf{67.49} & \textbf{75.82} & \textbf{80.95} & \textbf{84.29} & \textbf{62.99} \\
\bottomrule
\end{tabular}
\label{tab:capability_difficulty}
\end{table*}

\subsubsection{Stage Order Matters}
\label{sec:stage_order}

Table~\ref{tab:stage_order} analyzes the impact of different stage orders on visual math and perception performance. Across both Qwen2.5-VL-7B and Qwen3-VL-8B, we observe that the order of staged training plays a critical role in determining final model performance.
For both model series, Stage~1$\rightarrow$2$\rightarrow$3 (visual perception $\rightarrow$ textual reasoning $\rightarrow$ visual reasoning) consistently yields strong and balanced performance across visual math and perception benchmarks. Exchanging the first two stages (Stage~2$\rightarrow$1$\rightarrow$3) results in comparable average scores for math and perception. For the Qwen2.5-VL-7B model, these two training orders achieve 42.3\% \textit{v.s.} 42.9\% average scores across visual math benchmarks and 77.2\% \textit{v.s.} 76.3\% on visual perception, suggesting that visual perception and textual reasoning function as complementary foundational capabilities that can be learned in either order before visual reasoning.

In contrast, reversing the order to Stage~3$\rightarrow$2$\rightarrow$1 leads to a clear degradation in performance. For Qwen2.5-VL-7B, the visual math average score drops from over 42\% to 37.7\%, and the visual perception average decreases to 74.2\%, approaching the base model level. A similar trend is observed for Qwen3-VL-8B, where Stage~3$\rightarrow$2$\rightarrow$1 underperforms both Stage~1$\rightarrow$2$\rightarrow$3 and Stage~2$\rightarrow$1$\rightarrow$3 in overall accuracy (64.8\% \textit{v.s.} 65.8\% \textit{v.s.} 65.8\%). This indicates that prematurely training visual reasoning entangles perception and reasoning before either capability is sufficiently established.

Taken together, these findings indicate that staged training is not just about isolating different capabilities but also about acquiring them in a suitable sequence. Visual perception, as a fundamental skill, should be solidified before visual reasoning to maximize the effectiveness of staged training.

\subsection{RLVR is More Effective than SFT for Perception Training}
\label{sec:sft_vs_rl}

Caption-based supervised fine-tuning (SFT) is a widely adopted approach for aligning LLMs to the vision modality~\cite{liu2024improvedbaselinesvisualinstruction,chen2024allavaharnessinggpt4vsynthesizeddata,chen2023sharegpt4vimprovinglargemultimodal,ogezi2025spare,sun2024descriptive}, as it provides direct supervision on image-text correspondence. 
To examine whether this approach is suitable for enhancing perception at the post-training stage, we compare caption-based SFT with our RLVR approach in Stage 1 (visual perception) training, followed by the same training setups in subsequent stages.

As shown in Table~\ref{tab:rl_vs_sft}, RLVR consistently outperforms SFT across both Qwen2.5-VL-7B and Qwen3-VL-8B. In particular, RLVR leads to higher average visual perception scores (e.g., 77.2\% \textit{v.s.} 75.7\% on Qwen2.5-VL-7B and 80.4\% \textit{v.s.} 79.2\% on Qwen3-VL-8B), and these improvements translate into stronger visual math performance.
Notably, employing RLVR in visual perception training leads to an 8.2\% performance gain on Qwen2.5-VL-7B and the WeMath benchmark, increasing the average visual math score from 37.0\% to 42.3\%.
While SFT occasionally achieves competitive results on individual benchmarks (e.g., MathVision), RLVR provides more stable and consistent gains across both perception and reasoning metrics.
These results suggest that although caption-based SFT has been proven effective for vision-language alignment, RLVR offers a stronger training signal for perception by explicitly penalizing unsupported or hallucinated visual interpretations. As a result, RL-based perception training leads to more reliable visual grounding and improved downstream reasoning performance.

\subsection{Complementarity with Difficulty-Based Curriculum}
\label{sec:capability_difficulty}

Our staged training can be viewed as a \emph{capability-based curriculum}---organizing training by functional role (perception $\rightarrow$ reasoning) rather than by sample difficulty. To investigate whether this new curriculum dimension is complementary to traditional difficulty-based ordering, we compare four training configurations on Qwen3-VL-8B: merged training (no curriculum), capability-only (our staged training), difficulty-only (samples ordered by hardness within merged training), and the combination of both (difficulty ordering applied within each capability stage).
To estimate sample difficulty, we sample 16 answers per question from Qwen3-VL-8B with temperature 1.0 and compute the average pass rate as a difficulty score. Training samples are then ranked from easy (high pass rate) to hard (low pass rate). For \emph{difficulty-only}, we apply this ranking to the entire merged dataset; for \emph{capability+difficulty}, we apply the ranking \emph{within} each of the three capability stages and train the stages in our standard order (perception $\rightarrow$ textual reasoning $\rightarrow$ visual reasoning), with easy samples preceding hard ones in every stage. We evaluate on a diverse set including MathVerse Vision Only subset~\citep[MVerse (VO);][]{zhang2024mathverse}, DynaMath~\cite{zou2024dynamath}, CV-Bench~\cite{tong2024cvbench}, and V*Bench~\cite{wu2024vstarbench}.

As shown in Table~\ref{tab:capability_difficulty}, both capability-based and difficulty-based curricula individually improve over merged training (60.53\% and 60.36\% \textit{v.s.} 58.56\%). Crucially, combining the two dimensions yields a further substantial gain to 62.99\%, surpassing either curriculum alone by over 2\%. This demonstrates that capability-based staging and difficulty-based ordering address orthogonal aspects of training optimization and can be effectively composed for additive improvements.

%% file: sections/conclusion.tex
\section{Discussion and Conclusion}

In this work, we establish that visual perception is a dominant limiting factor for visual reasoning in VLMs and that longer reasoning alone cannot compensate for perceptual errors. Motivated by this insight, we introduce a staged post-training paradigm that decouples VLM capabilities into visual perception, textual reasoning, and visual reasoning stages. This decoupled approach consistently outperforms unified training pipelines across four model architectures while producing shorter reasoning traces, and we demonstrate that RLVR provides a more effective training signal than caption-based SFT for perception optimization.

Conceptually, our staged framework can be viewed as \emph{capability-dimension curriculum learning}---a framework that complements existing difficulty-dimension curricula~\cite{zhang2025learning,liu2024let}. Rather than scaling tasks by difficulty, we structure training by functional roles, and show that combining both curriculum dimensions yields further additive improvements (Section~\ref{sec:capability_difficulty}). This suggests a promising direction for multidimensional training trajectories in future VLM post-training.

\paragraph{Limitations.}
Our study has several limitations. First, all experiments are conducted at the 7--8B parameter scale; validation on larger models (32B+) remains future work. Second, our perception data pipeline relies on the availability of fine-grained image captions, which may limit applicability to domains without such resources. Third, our three-stage decomposition may not represent the finest granularity of capability separation; exploring more fine-grained stage decompositions is an interesting direction.